\documentclass[runningheads]{llncs}
\usepackage{graphicx}
\usepackage{amsfonts}
\usepackage{booktabs}
\usepackage{siunitx}

\usepackage[center]{caption}

\usepackage{amsmath}

\usepackage{xcolor}
\usepackage{tikz}
\usepackage{arydshln}
\usepackage{pgfplots}
\usepackage{verbatim}
\usetikzlibrary{arrows, automata}
\usetikzlibrary{positioning}
\usetikzlibrary{arrows.meta}
\newcommand\copyrighttext{%
  D. Bacciu and L. Di Sotto, A Non-Negative Factorization approach to node pooling in Graph Convolutional Neural Networks \emph{To appear in the Proceedings of the 18th International Conference of the Italian Association for Artificial Intelligence (AIIA 2019)}, Springer, 2019}
\newcommand\copyrightnotice{%
\begin{tikzpicture}[remember picture,overlay]
\node[anchor=north,yshift=-10pt] at (current page.north) {\fbox{\parbox{\dimexpr\textwidth-\fboxsep-\fboxrule\relax}{\copyrighttext}}};
\end{tikzpicture}%
}

\begin{document}
\title{A Non-Negative Factorization approach to node pooling in Graph Convolutional Neural Networks}
%
%
\author{Davide Bacciu \and
Luigi Di Sotto }
\authorrunning{D. Bacciu and L. Di Sotto}
\titlerunning{A NMF approach to node pooling in GCNs}
%
\institute{Universit\`a di Pisa\\
Dipartimento di Informatica \\
Largo B. Pontecorvo, 3 - Pisa (Italy) \\
\email{bacciu@di.unipi.it}\\
\email{l.disotto@gmail.com}}

\maketitle              

\copyrightnotice

\begin{abstract}
The paper discusses a pooling mechanism to induce subsampling in graph structured data and introduces it as a component of a graph convolutional neural network. The pooling mechanism builds on the Non-Negative Matrix Factorization (NMF) of a matrix representing node adjacency and node similarity as adaptively obtained through the vertices embedding learned by the model. Such mechanism is applied to obtain an incrementally coarser graph where nodes are adaptively pooled into communities based on the outcomes of the non-negative factorization. The empirical analysis on graph classification benchmarks shows how such coarsening process yields significant improvements in the predictive performance of the model with respect to its non-pooled counterpart.

\keywords{Graph Convolutional Neural Networks  \and Differentiable Graph Pooling \and Non-Negative Matrix Factorization.}
\end{abstract}
\section{Introduction}
Nowadays many real-world phenomena are modeled as interacting objects possibly living into high-dimensional manifolds with added topological structure. Examples can be found in genomics with protein-protein interaction networks, fake news discovery in social networks, functional networks in neuroscience. Graphs are the natural mathematical model for such data with underlying non-Euclidean nature.
Current Euclidean Convolutional Neural Networks have built their success leveraging on the statistical properties of stationarity, locality and compositionality of flat domains. Rendering convolutional neural networks able also to learn over non-Euclidean domains is not that straightforward in that is required a re-designing of the computational model for adaptively learning graph embeddings.
Over flat domains, i.e. grid-like structures, convolutional filters are compactly supported because of the grid regularity and the availability of consistent node ordering across different samples. This makes it possible to learn filters of fixed size and independent of the input signal dimension leveraging, to this end, weight sharing techniques. Furthermore, a set of symmetric functions is also applied for sub-sampling purposes to fully exploit the multi-scale nature of the grids.
The same does not apply to domains with highly varying topologies where learnt filters (non-Toeplitz operators) may be too representative of the considered domain, since they highly depend on the eigen-basis of the filter operator and they may thus fail to model sharp changes in the graph signal.
State-of-the-art Graph Convolutional Networks (GCNs) \cite{DBLP:journals/corr/DefferrardBV16,DBLP:journals/corr/KipfW16} try to overcome the above difficulties with convolutions based on $k$-order Chebyshev polynomials, introducing the interesting duality of implicitly learning the graph spectrum by simply acting on the spatial representation. GCNs efficiently avoid the computational burden of performing a spectral decomposition of the graph, yielding to learned filters that are independent of the number of nodes in the graph.  When considering graph classification tasks, we lack a principled multi-resolution operator providing coarser and more abstract representations of the input data as we go deeper in the network. Standard approaches to graph pooling employ symmetric functions such as max, summation or average along features axes of the graph embeddings.
In \cite{DBLP:journals/corr/abs-1810-00826}, it is given an account of the discriminative power of these different coarsening operators. In the present work, we introduce a simple pooling operator for graphs that builds on the Non-Negative Matrix Factorization (NMF) methods to leverage on the community structure underlying graph structured data to induce subsampling, or equivalently, a multiscale view of the input graph in order to capture long-range interactions as we go deeper in Graph Convolutional Networks (GCNs).
That would be of practical interest especially in the context of graph classification or regression tasks where the whole graph is fed into downstream learning systems as a single signature vector.
Such mechanism is thus applied to incrementally obtain coarser graphs where nodes are pooled into communities based on the soft assignments output of the NMF of the graph adjacency matrix and Gram matrix of learned graph embeddings.
Results on graph classification tasks show how jointly using such a coarsening operator with GCNs translate into improved predictive performances.

\section{Background}

In the following we introduce some basic notation used throughout the paper, then we briefly introduce the necessary background to understand state-of-the-art Graph Convolutional Neural Networks (GCNs). We mainly refer to spectral graph theory as introduced in \cite{Belkin:2001:LES:2980539.2980616,DBLP:journals/corr/BronsteinBLSV16,bruna}.

\subsection{Basic notation} A graph $G$ is a tuple $G = \left(\mathcal{V}, \mathcal{E}\right)$, where $\mathcal{V}$ is the set of vertices of the graph and $\mathcal{E}$ is the set of edges connecting vertices, i.e. $\mathcal{E} \subseteq \mathcal{V}\times \mathcal{V}$. Let $\mathcal{N}(i)$ be the set of neighbours of a node $i \in \mathcal{V}$. And let $A\in \bbbr^{n\times n}$, with $n = |\mathcal{V}|$, be the adjacency matrix such that
$$
	\displaystyle A_{i,j} = \left\{
									\begin{array}{ll}
										a_{i,j}>0  & \mbox{if } (i,j)\in\mathcal{E} \\
										0 & \mbox{otherwise. }
								    \end{array}
								\right.
$$

Note that in the above formulation we consider undirected graphs, i.e. such that $(i,j)\in\mathcal{E}$ and $(j,i)\in\mathcal{E}$. Thus, matrix $A$ is such that $A = A^T$. In the present work, without loss of generality, we generalize to undirected graphs

We also indicate with $X \in \bbbr^{n\times d}$ as the matrix of the $n$ signals $x_i\in \bbbr^d$ associated to each node $i \in \mathcal{V}$. 

\subsection{Graph Convolution via polynomial filters}

\subsubsection{Spectral construction.}
A first approach to representation learning on graphs is to explicitly learn the graph spectrum. In matrix notation, we can express the
generalized convolution over graphs as follows \cite{DBLP:journals/corr/BronsteinBLSV16}
\begin{equation} \label{spectral_graph_conv}
	L X = U \Lambda U^T X
\end{equation}
where $L$ is the combinatorial graph Laplacian, $L = D - A$, with $D$ the degree matrix such that $D_{ii} = \sum_{j} a_{ij}$, where $U \in \bbbr^{n\times k}$ is an orthonormal basis generalizing the Fourier basis, and where $\Lambda$ is a diagonal matrix being the spectral representation of the filter \cite{Belkin:2001:LES:2980539.2980616,bruna}. Matrices $U$ and $\Lambda$ are the solution to the generalized eigenvalue problem $L U = U \Lambda$ \cite{Belkin:2001:LES:2980539.2980616,bruna}.
With such an approach there are multiple problems: (a) the eigendecomposition in (\ref{spectral_graph_conv}), and its application (filtering), require non-trivial computational time; (b) the corresponding filters are non-localized \cite{DBLP:journals/corr/DefferrardBV16}; (c) filter size is $O(n)$, hence introducing a direct link between the parameters and the $n$ nodes in the graph (no weight sharing).

\subsubsection{Spatial construction.}
In \cite{DBLP:journals/corr/DefferrardBV16}, it is proposed an alternative approach to explicit learning of the graph spectrum, by showing how it can be learned implicitly through a polynomial expansion of the diagonal operator $\Lambda$. Formally,
\begin{equation}
	g_{\theta}\left( \Lambda \right) = \sum_{k=0}^{K-1} \theta_{k} \Lambda^{k}
\end{equation}
where $\theta \in \bbbr^{K}$ is the vector of polynomial coefficients. In \cite{DBLP:journals/corr/DefferrardBV16} is pointed out that spectral filters represented as $K$-order polynomials are exactly K-localized and that weight sharing is thus made possible, since filters have size $O(K)$. Graph CNN (GCNN), also known as ChebNet \cite{DBLP:journals/corr/DefferrardBV16}, exploited the previous observation by employing Chebyshev polynomials for approximating filtering operation (\ref{spectral_graph_conv}). Chebyshev polynomials are recursively defined using the recurrence relation
\begin{equation}\label{cheby-poly}
	\begin{split}
		T_{j} (\lambda) & = 2 \lambda T_{j-1} (\lambda) - T_{j-2} (\lambda); \\ T_{0} (\lambda) &= 1; \\ T_{1} (\lambda) &= \lambda.
	\end{split}
\end{equation}
Also, polynomials recursively generated by (\ref{cheby-poly}) form an orthonormal basis in $\left[-1, 1\right]$ \cite{DBLP:journals/corr/BronsteinBLSV16,DBLP:journals/corr/DefferrardBV16}. A filter can thus be represented as a polynomial of the form
\begin{equation}
	\begin{split}
  g_{\theta} (\hat{L}) & = \sum_{k=0}^{K-1} \theta_k U T_{k}(\hat{\Lambda}) U^{T} \\ & = \sum_{k=0}^{K-1} \theta_{k} T_{k} (\hat{L}),
  	\end{split}
\end{equation}
where $\hat{L} = 2 \Lambda / \lambda_{\max} - I_{n}$ and $\hat{\Lambda} = 2 \Lambda / \lambda_{\max} - I_{n}$ indicate a rescaling of the Laplacian eigenvalues to $\left[-1, 1\right]$. The filtering operation in (\ref{spectral_graph_conv}) can be rewritten, for one-dimensional input graph signals, as $\hat{x} = g_{\theta} (\hat{L}) x \in \bbbr^{n}$, where the $k$-th polynomial $\hat{x}_{k} = T_{k} (\hat{L}) x$ can be computed using the recurrence relation in (\ref{cheby-poly}) now defined as $\hat{x} = 2 \hat{L} x_{k-1} - \hat{x}_{k-2}$ with $\hat{x}_{0} = x$ and $\hat{x}_{1} = \hat{L} x$. More generally, taking into account multi-dimensionality of input data, we have a convolutional layer as follows
\begin{equation}\label{chebnet}
	\hat{X} = \sigma \left( \sum_{k=0}^{K-1} T_k\left(\Delta\right) X \Theta_k \right)
\end{equation}
with $\sigma$ a non-linear activation, and $\Theta \in \bbbr^{d_{in} \times d_{out}}$ the matrix of learnable parameters, with $d_{in}$ number of input features and $d_{out}$ number of neurons. A widely used convolutional layer over graphs are GCNs by \cite{DBLP:journals/corr/KipfW16} that are layers of the form of (\ref{chebnet}) with $K = 2$, namely
\begin{equation}\label{gcn}
	\hat{X} = ReLU \left( \hat{A}X\Theta \right).
\end{equation}

The $\Theta$ term, the matrix of polynomial coefficients to be learned, stems from (\ref{chebnet}) by imposing $\Theta_0 = - \Theta_1$, and with $\hat{A} = A + I$, and non-linearity being the ReLU function \cite{DBLP:journals/corr/KipfW16}. 
Thus, the main idea is to generate a representation for a node $i \in \mathcal{V}$ by aggregating its own features $x_{i} \in \bbbr^{d}$ and its neighbors’ features $x_{j} \in \bbbr^{d}$, where $j \in \mathcal{N}(i)$.
Note that, apart from the formulation meant to highlight the symmetry with convolutions on image data, the GCN model is not substantially different from the contextual approach to graph processing put forward by \cite{micheli} a decade before GCN, and recently extended to a probabilistic formulation \cite{pmlr-v80-bacciu18a} by leveraging an hidden tree Markov model \cite{gtmsd} with relaxed causality assumptions and a fingerprinting approach to structure embedding \cite{tnnlsKernel}.

\subsection{Node Pooling in Graph CNNs}

A first attempt to formalize graph pooling can be found in \cite{bruna}, a simple framework for multiresolution clustering of a graph is given based on a naive agglomerative method. There are some recent works proposing pooling mechanisms for graph coarsening in Deep GCNs, in \cite{2018arXiv181101287C} a subset of the nodes are dropped based on a learnable projection vector where at each layer only the top-$k$ interesting nodes are retained.
In \cite{DBLP:journals/corr/HamiltonYL17}, it is employed a rough node sampling and a differentiable approach through a LSTM model for learning aggregated node embeddings, though it may render difficult satisfying invariance with respect to node ordering.
Interestingly, in \cite{DBLP:journals/corr/abs-1901-01343} it is applied a simple and well known method from Graph Theory for node decimation based on the largest eigenvector $u_{max}$ of the graph Laplacian matrix. 
They further employ a more sophisticated procedure to reduce Laplacian matrix using the sparsified Kron reduction.
Another relevant differentiable approach is that
put forward by DiffPool \cite{NIPS2018_7729}, where the model learns soft assignments to pool similar activating patterns into the same cluster, though the idea of learning hiearchical soft-clustering of graphs via adjacency matrix decomposition using a symmetric variant of NMF can be dated back to \cite{NIPS2005_2948}. In DiffPool, the learned soft assignment matrix is applied as a linear reduction operator on the adjacency matrix and the input signal matrix, and the coarsened graph is thus further convolved with GCNs.

\section{NMFPool: node pooling by Non-Negative Matrix Factorization}

In the following section we introduce our model, NMFPool, a principled Pooling operator enabling deep graph CNNs develop multi-resolution representations of input graphs. NMFPool leverages community structure underlying graphs to pool similar nodes to progressively gain coarser views of a graph. To that end we take inspiration from \cite{NIPS2005_2948} in which latent community structure of graph data is made explicit via adjacency matrix decomposition using Symmetric NMF (SNMF). NMFPool is grounded on that idea, building, instead, on a general non-symmetrical NMF of the adjacency matrix without constraining solutions to be stochastic. Before going further into details of our approach, we first introduce the formal definition of the NMF problem, then we give an intuitive interpretation of its solutions to clarify why NMF would help solve the graph pooling problem on graphs. At the end we will show how to use product factors of NMF as linear operators to aggregate topology and content information associated to graphs.
NMF is a popular technique for extracting salient features in data by extracting a latent space representation of the original information. Throughout the paper we refer to the original idea of NMF \cite{NIPS2000_1861} though it has been extensively studied in numerical linear algebra in the last years by many authors and for a variety of applications. Formally, the NMF problem can be stated as follows:
\begin{definition}
	Given a non-negative matrix $A \in \bbbr^{n \times m}_{+}$, find non-negative matrix factors $W\in\bbbr^{n \times k}_{+}$ and $H\in\bbbr^{k \times m}_{+}$, with $k < \min(m, n)$, such that
\end{definition}
\begin{equation}\label{nmf-problem}
	A \approx W H
\end{equation}

If we see matrix $A$ as having $m$ multivariate objects column-stacked, the straightforward interpretation of (\ref{nmf-problem}) is as follows
\begin{equation}\label{nmf-intepretation}
	a_j \approx W h_j,
\end{equation}
with $a_j$ and $h_j$ corresponding to $j$-th columns of $A$ and $H$.  The approximation (\ref{nmf-intepretation}) entails that each multi-variate object is a linear combination of columns of $W$ weighted by coefficients in $h_j$. Thus $W$ is referred to as the basis matrix or equivalently the cluster centroids matrix if we intend to interpret NMF as a clustering method. Matrix $H$ can be seen, instead, as a low-dimensional representation of the input data making thus NMF also useful for dimensionality reduction. Latent representation, in the clustering perspective, may indicate whether a sample object belongs to a cluster. For example, we could constrain each data-point to belong to a single cluster at a time: namely, each data-point is assigned to the closest cluster $x_j \approx u_j$. We generally look for non-trivial encodings to explain community evolution in graphs. Thus, the problem could be relaxed to a soft-clustering problem in that each data-point can belong to $k$ overlapping clusters \cite{Watt:2016:MLR:3126125}.  Formulation (\ref{nmf-problem}) requires to define a metric to measure the quality of the approximation, and  Kullback-Leibler (KL-) divergence or the more common Frobenius norm (F-norm) are common choices.
Many techniques from numerical linear algebra can be used to minimize problem (\ref{nmf-problem}) whatever the cost function we use, although its inherently non-convex nature does not give any guarantee on global minimum \cite{2019arXiv190301321F}. In  \cite{NIPS2000_1861} were first proposed multiplicative and additive update rules that ensure monotone descrease under KL- or F-norm.

\begin{figure}
\centering
\includegraphics[width=.8 \textwidth]{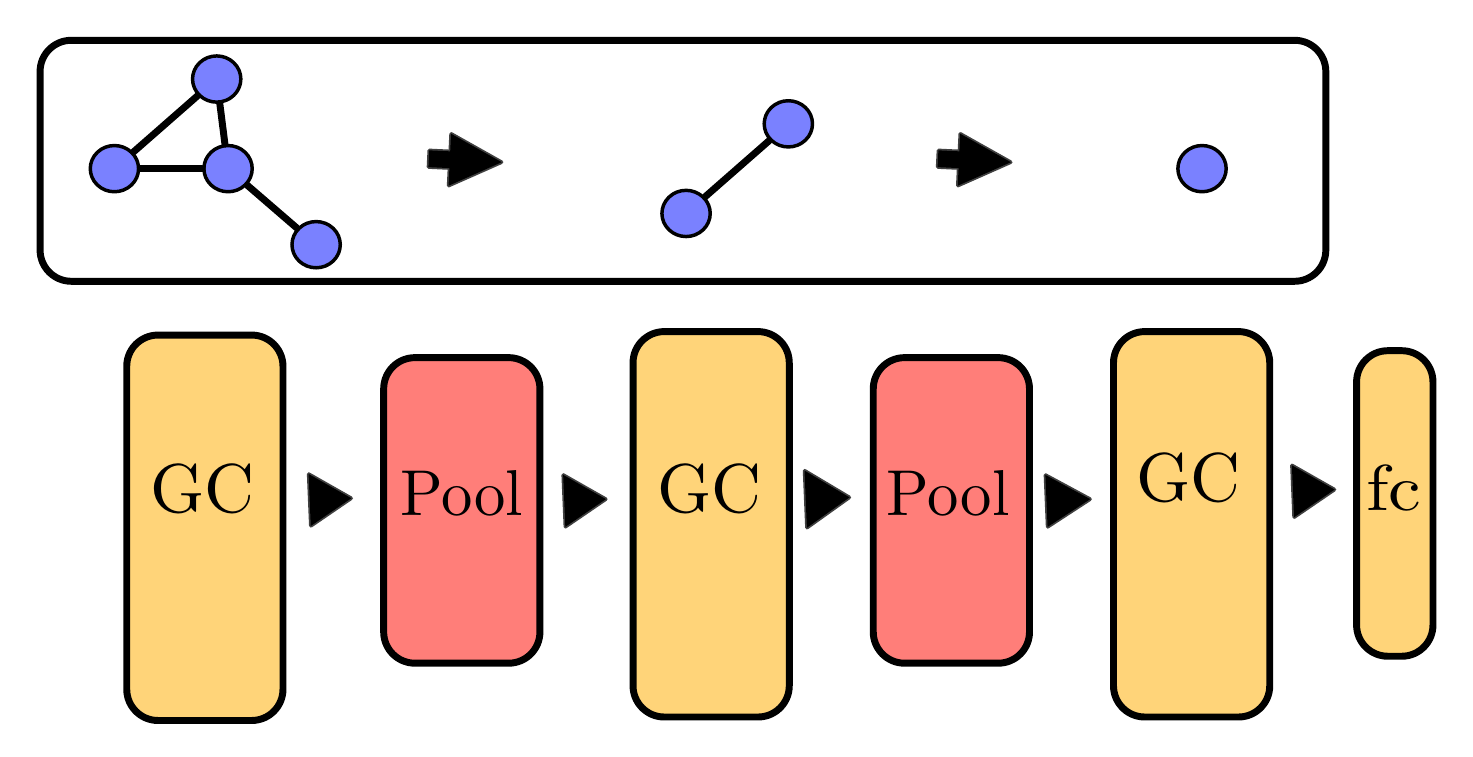}
\caption{High level architecture of a 3-layers GCN interleaved with 2 NMF Pooling layers.} \label{figArch}
\end{figure}

Thus, our proposed solution can be summarized into two main steps. First, we encode the input adjacency matrix to learn soft-assignments of nodes, and that could accomplished via exact NMF of the adjacency matrix. Second, we apply soft-assignments as linear operators to coarse adjacency matrix and node embeddings.  To this end, we refer to algebraic operations seen in \cite{NIPS2005_2948} for decomposing adjacency matrices and we extend it using equations widely used for graph coarsening \cite{NIPS2018_7729}, for they take into account embedding matrix reduction and nodes connectivity strength. For a complete picture, consider $\ell$ NMFPool layers interleaved with at least $\ell + 1$ stacked Graph Convolutions (GCs) as illustrated in Figure \ref{figArch}, where the graph convolutions are computed according to (\ref{gcn}). Then, let $Z^{(i)} \in \bbbr^{n_{i}\times d}$ be the output of $i$-th GC, namely the convolved node embeddings at layer $i$-th, defined as
\begin{equation}\label{gc-embed}
	Z^{(i)} = \mbox{ReLU}\left( A^{(i)}Z^{(i-1)}\Theta^{(i)} \right)
\end{equation}
with adjacency matrix $A^{(i)}\in\bbbr^{n_i \times n_i}$, with $n_i$ number of nodes at previous layer, and $\Theta^{(i)} \in \bbbr^{d \times d}$ matrix of weights. Observe that we are assuming, without loss of generality, each GC layer (\ref{gc-embed}) as having the same number of neurons. Observe also that $Z^{(0)} = X \in \bbbr^{n\times d}$, namely the initial node labels, and the initial adjacency matrix is set to $A^{(0)} = \hat{D}^{-1/2} \hat{A} \hat{D}^{-1/2}$, i.e. the normalized adjacency matrix with $\hat{A} = A + I$, $A \in \bbbr^{n\times n}$, and $\hat{D}$ is a diagonal matrix of node degrees \cite{DBLP:journals/corr/KipfW16}.

The $i$-th NMFPool layer solves the problem in (\ref{nmf-problem}), i.e. the decomposition of the symmetric and positive $A^{(i)}$, by minimizing the following loss
\begin{equation}\label{nmf-loss}
	|| A^{(i)} - W^{(i)} H^{(i)} ||_{F}
\end{equation}
with $W^{(i)} \in \bbbr_{+}^{n_i\times k_i}$ and $H^{(i)} \in \bbbr_{+}^{k_i \times n_i}$, and $k_i$ number of overlapping communities to pool the $n_i$ nodes into, and $\| . \|_F$ the Frobenius norm. Observe that $k_i$'s are  hyper-parameters to control graph coarsening scale. The algorithm to minimize (\ref{nmf-loss}) depends on the underlying NMF implementation. Then NMFPool applies the encoding $H^{(i)}$ to coarsen graph topology and its content as follows
\begin{equation}\label{nmfpool-application-a}
Z^{(i+1)} = H^{(i)T} Z^{(i)} \in \bbbr^{k_i \times d}
\end{equation}
\begin{equation}\label{nmfpool-application-b}
	A^{(i+1)} = H^{(i)T} A^{(i)} H^{(i)} \in \bbbr^{k_i \times k_i}.
\end{equation}

\begin{figure}
\centering
\includegraphics[width=.8 \textwidth]{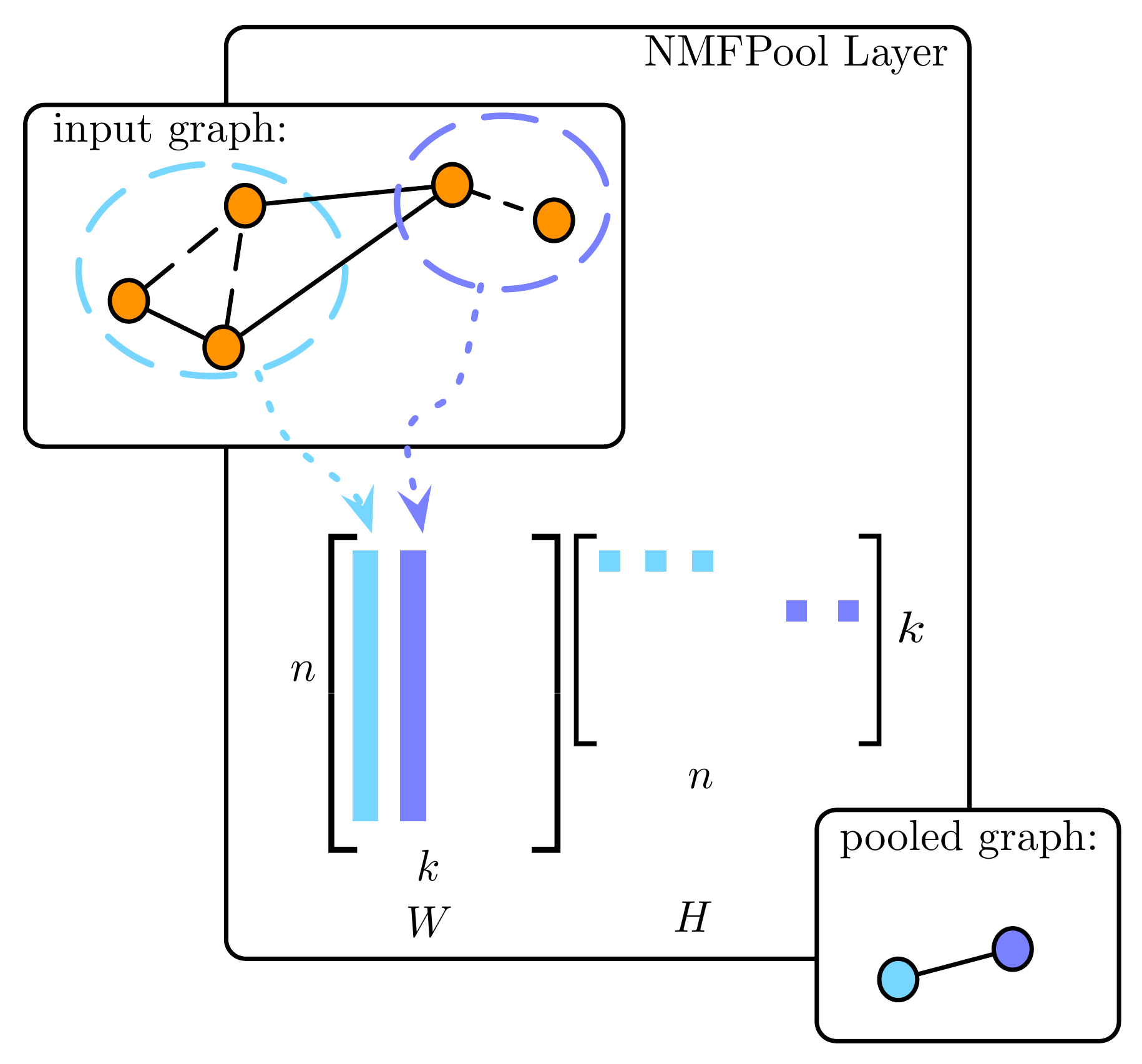}
\caption{The NMFPool layer. Orange circles represent nodes of input graph, and solid lines the edges. Dashed lines are the predicted edges in between nodes pooled together. Colored dashed circles represent discovered communities.} \label{figPool}
\end{figure}

A graphical interpretation of the inner workings of the NMFPool layer is provided in Figure \ref{figPool}, highlighting the interpretation of pooling as a matrix decomposition operator. It is crucial to point out that NMFPool layers are independent of the number of nodes in the graph, which is essential to deal with graphs with varying topologies.

\section{Experiments}

We assess the effectiveness of using the exact NMF of the adjacency matrix $A$ as a pooling mechanism in graph convolutional neural networks. To this end, we consider five popular graph classification benchmarks and we further compare the performance of our approach, referred to as NMFPool in the following,  with that of DiffPool, with the goal of showing how a simple and general method may easily compare to differentiable and parameterized pooling operators such as DiffPool. Results were gathered on graph classification tasks for solving biological problems on the ENZYMES (\cite{borgwardt2005protein}, \cite{brenda_enzymes}),  NCI1 \cite{Wale:2008:CDS:1357641.1357642}, PROTEINS (\cite{borgwardt2005protein}, \cite{DOBSON2003771}), and D\&D (\cite{DOBSON2003771}, \cite{Shervashidze:2011:WGK:1953048.2078187}) datasets and the scientific collaboration dataset COLLAB \cite{Yanardag:2015:DGK:2783258.2783417}. In Table \ref{table-datasets} are summarized statistics on benchmark datasets.

\begin{table}
\centering
\caption{Statistics on benchmark datasets.}\label{table-datasets}
\begin{tabular}{ccccc} \toprule[1.5pt]
    {\bfseries{Dataset}} & {\bfseries{Graphs}} & {\bfseries{Classes}} & {\bfseries{Nodes (avg)}} & {\bfseries{Edges (avg)}} \\ \midrule
    {COLLAB} & 5000 & 3 & 74.49 & 2457.78 \\
    {D\&D} & 1178 & 2 & 284.32 & 715.66 \\
    {ENZYMES} & 600  & 6 & 32.63 & 62.14   \\
    {NCI1} & 4110 & 2 & 29.87 & 32.30 \\
    {PROTEINS} & 	1113 & 2 & 39.06 & 72.82 \\ \bottomrule[1.5pt]

\end{tabular}
\end{table}

In our experiments, the baseline graph convolution is the vanilla layer in (\ref{gcn}). For both models, we employed the interleaving of pooling and convolutional layers depicted in the architecture in Figure \ref{figArch}, varying the number of pooling-convolution layer pairs to assess the effect of network depth on task performance. Note that the number of layers in the convolutional architecture influences the context spreading across the nodes in the graph. Implementation of NMFPool and Diffpool is based on the Pytorch Geometric library \cite{DBLP:journals/corr/abs-1903-02428}, complemented by the NMF implementation available in the Scikit library. Models configurations were run on a multi-core architecture equipped with 4 NUMA nodes each with 18 cores (Intel(R) Xeon(R) Gold 6140M @ 2.30GHz) capable of running 2 threads each for a total of 144 processing units available. We had access also to 4 Tesla V100 GPUs accelerators.

Model selection was performed for exploring a variety of configurations using stratified 3-fold cross validation. Following standard practice in graph convolution neural networks, learning rate was set with an initial value of $0.1$ and then decreased by a factor of $0.1$ whenever validation error did not show any improvement after $10$ epochs wait. The number of neurons is the same for each graph convolutional layer and it has been selected in $\{16, 32, 64, 128\}$ as part of the cross-validation procedure. When applying the pooling operator both NMFPool and Diffpool require to define the number of communities $k$, similarly to how the pooling operator on images requires the definition of the pooling windows size (and stride). Here, following the idea indicated in the original DiffPool paper  \cite{NIPS2018_7729},  we choose different $k$ for each dataset as a fraction of the average number of nodes in the samples. Thus during cross-validation we intended to study how NMFPool and DiffPool behave as a function of the cluster sizes $k_i$ at each layer. To this end, pooling size has been selected from the set $\{k_1, k_2\}$. In particular, for models with a single pooling layer, we tested both sizes $k_1$ and $k_2$.  Instead, for deeper architectures, we restricted to use the largest $k_i$ for the first layer, following up in decreasing order of $k_i$. Table \ref{table-fractions} summarizes the number of clusters used for the first and second pooling layer in the architectures considered in this empirical assessment.

\begin{table}
\centering
\caption{$k_1$ is computed using formula $k_1 = \lfloor n_{avg} \cdot p\rfloor $ with $p$ varying in $\left[21\% - 25\%\right]$, and $n_{avg}$ average number of nodes (see Table \ref{table-datasets}). Then $k_2 = k_1 / 2$. Fractions are chosen depending on the size of task at hand and to previous empirical observation. Except for the D\&D dataset where $p = 5\%, 1\%$, being the bigger dataset we needed a good compromise between abstraction capability and computational time.}\label{table-fractions}
\begin{tabular}{cccc} \toprule[1.5pt]
    {\bfseries{Dataset}} & {\bfseries{$k_1$}} & {\bfseries{$k_2$}} & {\bfseries{$p$}} \\ \midrule
    {COLLAB} & 16 & 8  & 22\% \\
    {D\&D} & 14 & 2  &  5\% - 1\% \\
    {ENZYMES} & 8 & 4 & 25\% \\
    {NCI1} & 6 & 3 & 24\% \\
    {PROTEINS} & 8 & 4 & 21\% \\ \bottomrule[1.5pt]
\end{tabular}
\end{table}

The outcome of the empirical assessment is summarized in Table \ref{table-acc}, where it is reported the mean classification accuracy of the different models averaged on the dataset folds. Table \ref{table-acc} reports results for a vanilla GCN (no pooling) and a varying number of graph convolution layers: results show how at most two layers are sufficient to guarantee good performances, while three layers are only required for the COLLAB dataset and a single layer network obtains the best performance on the NCI1 dataset. In the experiments we thus decided to employ at most three GCN layers, namely at most two NMF and DiffPool pooling layers. It is still evident how adding more convolutional and pooling layers does not always result into better performances. The analysis of the results for NMFPool shows how the addition of the simple NMF pooling allows a consistent increase of the classification accuracy with respect to the non-pooled model for all the benchmark datasets. Note how a single pooling layer is sufficient, on most datasets, to obtain the best results, confirming the fact that pooling allows to effectively fasten the process of context spreading between the nodes. When compared to DiffPool, our approach achieves accuracies which are only marginally lower than DiffPool on few datasets. This despite the fact that DiffPool employs a solution performing an task-specific parameterized decomposition of the graph, while our solution simply looks for quasi-symmetrical product matrices by knowing nothing of the underlying task.
\begin{table}
\caption{Mean and standard deviation (in brackets) of graph classification accuracies on the different benchmarks, for the vanilla GCN with $\ell$ convolutional layers ($\ell$-GC), for NMFPool and DiffPool with $\ell_p$ pooling layers and $\ell_p+1$ convolutional layers (i.e. $\ell_{p_1}$-NMFPool and $\ell_{p_2}$-DiffPool, respectively).}\label{table-acc}
\begin{tabular}{cccccc} \toprule[1.5pt]
    {\bfseries{Model}} & {\bfseries{ENZYMES}} & {\bfseries{NCI1}} & {\bfseries{PROTEINS}} & {\bfseries{D\&D}} & {\bfseries{COLLAB}} \\ \midrule
    {1-GC} &  0.222$~\left(0.023\right)$ & 0.625 $\left(0.014\right)$ & 0.713 $\left(0.019\right)$ & 0.681 $ \left(0.045\right)$ & 0.671 $\left(0.007\right)$ \\
    {2-GCs}  & 0.228 $\left(0.023\right)$  & 0.620 $\left(0.057\right)$ & 0.720 $\left(0.034\right)$ & 0.704 $ \left(0.048\right)$ & 0.678 $\left(0.007\right)$   \\
    {3-GCs}  & 0.182 $\left(0.022\right)$ & 0.628 $\left(0.031\right)$ & 0.688 $\left(0.024\right)$  & 0.692 $\left(0.032\right)$ & 0.681 $\left(0.002\right)$ \\ \midrule
    {1-NMFPool}  & 0.241 $\left(0.039\right)$   & 0.662 $\left(0.026\right)$  & 0.721 $\left(0.031\right)$  & 0.760 $\left(0.015\right)$ & 0.650 $\left(0.004\right)$ \\
    {2-NMFPool}  & 0.175 $\left(0.023\right)$  & 0.655 $\left(0.013\right)$ & 0.724 $\left(0.020\right)$  & 0.753 $\left(0.010\right)$ & 0.658 $\left(0.002\right)$  \\ \midrule
    {1-DiffPool}  & 0.259 $\left(0.069\right)$   & 0.661 $\left(0.017\right)$  & 0.743 $\left(0.011\right)$ & 0.770 $\left(0.007\right)$ & 0.659 $\left(0.005\right)$ \\
    {2-DiffPool} & 0.239 $\left(0.064\right)$   & 0.632 $\left(0.017\right)$ & 0.744 $\left(0.026\right)$   & 0.761 $\left(0.003\right)$ & 0.667 $\left(0.022\right)$  \\ \bottomrule[1.5pt]
\end{tabular}
\end{table}

\section{Conclusions}
We introduced a pooling mechanism based on the NMF of the adjacency matrix of the graph, discussing how this approach can be used to yield a hierarchical soft-clustering of the nodes and to induce a coarsening of the graph structure.  We have empirically assessed our NMPool approach with the task-specific adaptive pooling mechanism put forward by the DiffPool model on a number of state-of-the-art graph classification benchmarks. We argue that our approach can yield to potentially more general and scalable
pooling mechanisms than DiffPool, allowing to choose weather the pooling mechanism has to consider the node embeddings computed by the model and the task-related information when performing the decomposition (as in DiffPool), but also allowing to directly decompose the graph structure a-priori with no knowledge of the node embeddings adaptively computed by the convolutional layer. This latter aspect, in particular, allows to pre-compute the graph decomposition and results in a multiresolution representation of the graph structure which does not change with the particular task at hand.

Future works will consider the use of symmetric and optimized NMF variants to increase prediction performances. It also would be of particular interest to improve the quality and quantity of information NMFPool retains into the encoding matrix. NMFPool could evolve out of its general purpose form, for example, making it a generative end-to-end differentiable layer using probabilistic approaches. See \cite{bayesian-nmf-lio} for an attempt to solve NMF using probabilistic models. We could refer to the popular probabilistic generative model of the Variational Auto-Encoders (VAEs) \cite{2013arXiv1312.6114K}, \cite{pmlr-v32-rezende14} possibly extended to graphs \cite{2016arXiv161107308K}. The underlying hierarchical structure of graph data may also be taken into account by imposing latent encoding to match priors referring to hyperbolic spaces \cite{2019arXiv190106033M}. Interestingly, latent matrix encoding may not be forced to match overimposed priors, for they could make the model too biased over particular graph geometries. Instead, such priors could be directly learned from relational data using adversarial approaches \cite{DBLP:journals/corr/MeschederNG17} extended also to graph auto-encoders \cite{DBLP:journals/corr/abs-1802-04407}. Another interesting feature would be to make NMFPool independent of hyper-parameter $k$.

\subsection*{Acknowledgments}
This work has been supported by the Italian Ministry of Education, University, and Research (MIUR) under project SIR 2014 LIST-IT (grant n. RBSI14STDE).
%
%
%

\end{document}